\documentclass[10pt,twocolumn,letterpaper]{article}

\usepackage{wacv}
\usepackage{times}
\usepackage{epsfig}
\usepackage{graphicx}
\usepackage{amsmath}
\usepackage{amssymb}

\usepackage{amsthm,url,balance}
\usepackage{color}
\usepackage[T1]{fontenc}

\usepackage{algorithm}
\usepackage{algorithmic}
\usepackage[ruled,vlined,algo2e]{algorithm2e}

\usepackage{dsfont}	
\usepackage{wrapfig}
\usepackage{footnote}
\usepackage{enumitem}

\newcommand{\bfsection}[1]{\vspace*{0.1cm}\noindent\textbf{#1.}}

\usepackage[pagebackref=true,breaklinks=true,letterpaper=true,colorlinks,bookmarks=false]{hyperref}

\wacvfinalcopy 

\def\wacvPaperID{322} 

\ifwacvfinal\fi
\begin{document}

\title{Towards Learning Affine-Invariant Representations via Data-Efficient CNNs}

\author{Wenju Xu\textsuperscript{1}\thanks{This work was done when the first author took an internship at MERL.}, Guanghui Wang\textsuperscript{2}, Alan Sullivan\textsuperscript{3}, Ziming Zhang\textsuperscript{3}\thanks{Corresponding author.} \\
\textsuperscript{1} Aerospace Engineering, The University of Kansas, Lawrence, KS 66045\\
\textsuperscript{2} Electrical Engineering and Computer Science, The University of Kansas, Lawrence, KS 66045\\
\textsuperscript{3} Mitsubishi Electric Research Laboratories (MERL), Cambridge, MA 02139\\
{\tt\small \{xuwenju, ghwang\}@ku.edu, \{sullivan, zzhang\}@merl.com}
}


\maketitle

\begin{abstract}

In this paper we propose integrating a priori knowledge into both design and training of convolutional neural networks (CNNs) to learn object representations that are invariant to affine transformations (\ie translation, scale, rotation). 
Accordingly we propose a novel multi-scale maxout CNN and train it end-to-end with a novel rotation-invariant regularizer. This regularizer aims to enforce the weights in each 2D spatial filter to approximate circular patterns. In this way, we manage to handle affine transformations in training using convolution, multi-scale maxout, and circular filters. Empirically we demonstrate that such knowledge can significantly improve the data-efficiency as well as generalization and robustness of learned models. For instance, on the Traffic Sign data set and trained with only 10 images per class, our method can achieve 84.15\% that outperforms the state-of-the-art by 29.80\% in terms of test accuracy.

\end{abstract}
	
\section{Introduction}
Recently Sabour \etal \cite{sabour2017dynamic} proposed a new network architecture, CapsNet, and a dynamic routing training algorithm to connect the capsules \cite{hinton2011transforming}, a new type of neurons that output vectors rather than scalars in conventional neurons, in two adjacent layers and group similar features in higher layers. Later on Hinton \etal \cite{hinton2018matrix} proposed another EM-based routing-by-agreement algorithm for training CapsNet. In contrast to CNNs, the intuition behind CapsNet is to achieve ``viewpoint invariance'' in recognizing objects for better generalization which is inspired by inverse graphics \cite{inverse-graphics}. Technically, CapsNet not only predicts classes but also encodes extra information such as geometry of objects, leading to richer representation. For instance, in \cite{hinton2018matrix}, $4\times 4$ pose matrices are estimated to capture the spatial relations between the detected parts and a whole. Unlike CNNs the performance of CapsNet on real and more complex data has not been verified yet, partially due to the high computation that prevents it from being applicable widely.



In fact exploring such invariant representations for object recognition has a long history in the literature of both neural science and computer vision. For instance, in \cite{isik2013dynamics} Isik \etal observed that object recognition in the human visual system is developed in stages with invariance to smaller transformations arising before invariance to larger transformations, which supports the design of feed-forward hierarchical models of invariant object recognition. In computer vision part-based representation (\eg \cite{fischler1973representation}) is one of the most popular invariant object representations that considers an object as a graph where each node represents an object part and each edge represents the (spatial) relation between the parts. Conceptually part-based representation is view-invariant in 3D and {\em affine-invariant} (\ie invariant to translation, scale, and rotation) in 2D. Although the complexity of part-based models in inference on general graphs could be very high \cite{crandall2005spatial}, for tree structures such as star graphs this complexity can be linear to the number of parts \cite{felzenszwalb2012distance}. Girshick \etal \cite{girshick2015deformable} has shown that such star graph part-based models can be interpreted as CNNs.


In this paper we aim to study the following problem: {\em Can we design and train CNNs to learn affine-invariant representations efficiently, effectively, and robustly?}

\bfsection{Motivation}
Besides CapsNet, we are also partially motivated by the works such as \cite{andreas2016neural, NIPS2017_6769} that utilize {\em a priori} knowledge as guidance to design and train neural networks efficiently and effectively. For instance, \cite{andreas2016neural} proposed the notion of ``neural module'' to conduct certain semantic functionality using deep learning for visual question answering. Such modules can be reusable to comprise complex networks to perform certain tasks. The semantics and the network design here come from the compositional linguistic structure of questions. Thanks to these modules, the network design is much more understandable by checking whether the outputs of a module follow what we expect. \cite{NIPS2017_6769} proposed encoding network weights as well as the architecture into a Tikhonov regularizer by lifting the ReLU activations, and accordingly developed a block coordinate descent algorithm for fast training of deep models.

In contrast to {\em a posteriori} knowledge such as visualization of learned filters in \cite{zhang2018interpretable}, a priori knowledge based approaches are more likely to be model-driven so that one can derive by reasoning alone, rather than being data-driven, 
in terms of building automatic systems such as neural networks. In this way, the networks with a priori knowledge are expected to be much easier to be understood by human, and their performance is more predictable and robust.



\bfsection{Contributions}
Thanks to the convolution, CNNs are translation equivariant. 
This capability has contributed significantly to their widespread success. They, however, are not efficient or effective to capture the scaled or rotated objects, and thus enhancing CNNs with the capability of learning scale-invariant and rotation-invariant features is very challenging but appealing. 

In this paper we design a novel deep multi-scale maxout \cite{goodfellow2013maxout} CNN to learn scale-invariant representations. We then propose training this network end-to-end with a novel rotation-invariant regularizer. 
To our best knowledge, we are the first to propose such regularization for handling rotation in deep learning. Note that we take the multi-scale maxout block and the regularizer as a priori knowledge for learning affine-invariant representations. Empirically we demonstrate the benefit of integrating such knowledge with network design and training, leading to better generalization, data-efficiency, and robustness of deep models than the state-of-the-art in learning affine-invariant representations. 


\section{Related Work}\label{sec:rw}

\bfsection{Scale-Invariant Networks} 
One simple way to handle the scale issue is using image pyramid in deep learning \cite{lin2017feature}. Some works \cite{xu2014scale,kanazawa2014locally,takahashi2017scale} are particularly interested in extracting scale-invariant features from the networks. More broadly, multi-scale convolutional filters (or multi-kernels) are employed in networks \cite{liao2015competitive,wang2016deeply,audebert2016semantic,larsson2016fractalnet}. The inception module in GoogLeNet \cite{szegedy2015going} is able to capture multi-scale information with maxout units. A similar idea has been explored in TI-Pooling \cite{laptev2016ti}. 
ResNet \cite{he2016deep} manages to capture multi-scale information using skip connection. Multi-scale DenseNet \cite{huang2017multi} proposes using a two-dimensional multi-scale convolutional network architecture that maintains coarse-level and fine-level features throughout the network. Note that with the increase of the number of hidden layers all the CNNs tend to extract deep features within multiple scales to a certain degree.

\bfsection{Rotation-Invariant Networks} 
Recently quite a few works focus on learning rotation-invariant features using deep networks. Cohen and Welling \cite{cohen2016group} proposed Group equivariant CNNs (GCNN) by exploiting larger groups of symmetries, including rotations and reflections, in the convolutional layers. Worrall \etal \cite{worrall2017harmonic} proposed Harmonic Networks by replacing regular CNN filters with circular harmonics and returning a maximal response and orientation for every receptive field patch. Both works argue that rotating the data point is equivalent to rotating the filters. Therefore, they manage to learn rotation-invariant filters in a continuous space. In contrast, some other works such as \cite{zhou2017oriented,zhang2017rotation,luan2018gabor,hoogeboom2018hexaconv,zhang2017interleaved,marcos2017rotation,weiler2018learning} propose learning the filters in a discretized space by quantizing the rotation angles with predefined numbers (\eg from 0 to $2\pi$, step by $\frac{\pi}{4}$) so that the final features encode the rotation information. For instance, Rotation Equivariant Vector Field Networks (RotEqNet) \cite{marcos2017rotation} was proposed by applying each convolutional filter at multiple orientations and returning a vector field that represents magnitude and angle of the highest scoring orientation at every spatial location. 

\bfsection{Interpretable Networks with A Priori Knowledge}
Andreas \etal \cite{andreas2016neural} proposed neural modules to mimic some basic semantic functionality using deep neural networks, based on which larger networks are constructed for specific tasks using the knowledge from natural language processing (NLP) such as grammar graphs as guidance. Belbute-Peres \etal \cite{de2018end} proposed embedding structured physics knowledge into larger systems as a differentiable physics engine that can be integrated as module in deep neural networks for end-to-end learning. Amos \etal \cite{amos2018differentiable} proposed using Model Predictive Control (MPC) as a differentiable policy class for reinforcement learning in continuous state and action spaces that leverages and combines the advantages of model-free and model-based approaches. They also showed that their MPC policies are significantly more data-efficient than a generic neural network.

\bfsection{Other Related Networks}
Dilated convolution \cite{yu2015multi} supports exponential expansion of the receptive field (\ie window) without loss of resolution or coverage and thus can help networks capture multi-scale information. Deformable Convolutional Networks (DCN) \cite{dai2017deformable} proposed a more flexible convolutional operator that introduces pixel-level deformation, estimated by another network, into 2D convolution. Spatial Transformer Networks (STN) \cite{jaderberg2015spatial} learn affine-invariant representations by sequential applications of a localization network, a parameterized grid generator and a sampler. Dynamic Filter Networks (DFN) \cite{jia2016dynamic,wu2018dynamic} was proposed to learn to generate (local) filters dynamically conditioned on an input that potentially can be affine-invariant.

\bfsection{Data Augmentation}
It is a well-known technique in deep learning for reducing the filter bias during learning by generating more (fake) data samples based on some predefined rules (or transformations) such as translation, scaling, rotation and random cropping. Trained with such augmented data, one can expect that the networks may be more robust to the transformations. For instance, TI-Pooling \cite{laptev2016ti} assembles all the transformed instances from the same data point in a pool and takes the maximal response for classification. STN \cite{jaderberg2015spatial} learns to predict a transformation matrix for each observation that can be used to augment data. 

\bfsection{Loss Functions}
From the perspective of the feature space, affine-invariant representations for an object under different transformations with translation, scale, and rotation should be mapped into a single point in the feature space ideally, or a compact cluster. To achieve this, several loss functions were proposed. For instance, the center loss \cite{wen2016discriminative} enforces the features from the same class to be close to the corresponding cluster center. Similar ideas have been explored in few-shot learning with neural networks \cite{snell2017prototypical} as well. In fact well-designed networks can generate compactly clustered features for each class with good discrimination, even if trained without such specific losses. Also such losses do not aim to learn affine-invariant features, explicitly or implicitly. Empirically we do not observe any improvement using the center loss over the cross-entropy loss, and thus we do not report the performance using the center loss.

In contrast to these previous works, we handle scale and rotation jointly in CNNs for learning affine-invariant representations. We introduce a priori knowledge into network design and training as interpretability in deep models. We demonstrate better generalization, data-efficiency, and robustness of our approach than the state-of-the-art networks.

\begin{figure}[t]
   \begin{minipage}{0.495\linewidth}
     \centering
     \includegraphics[width=.9\linewidth]{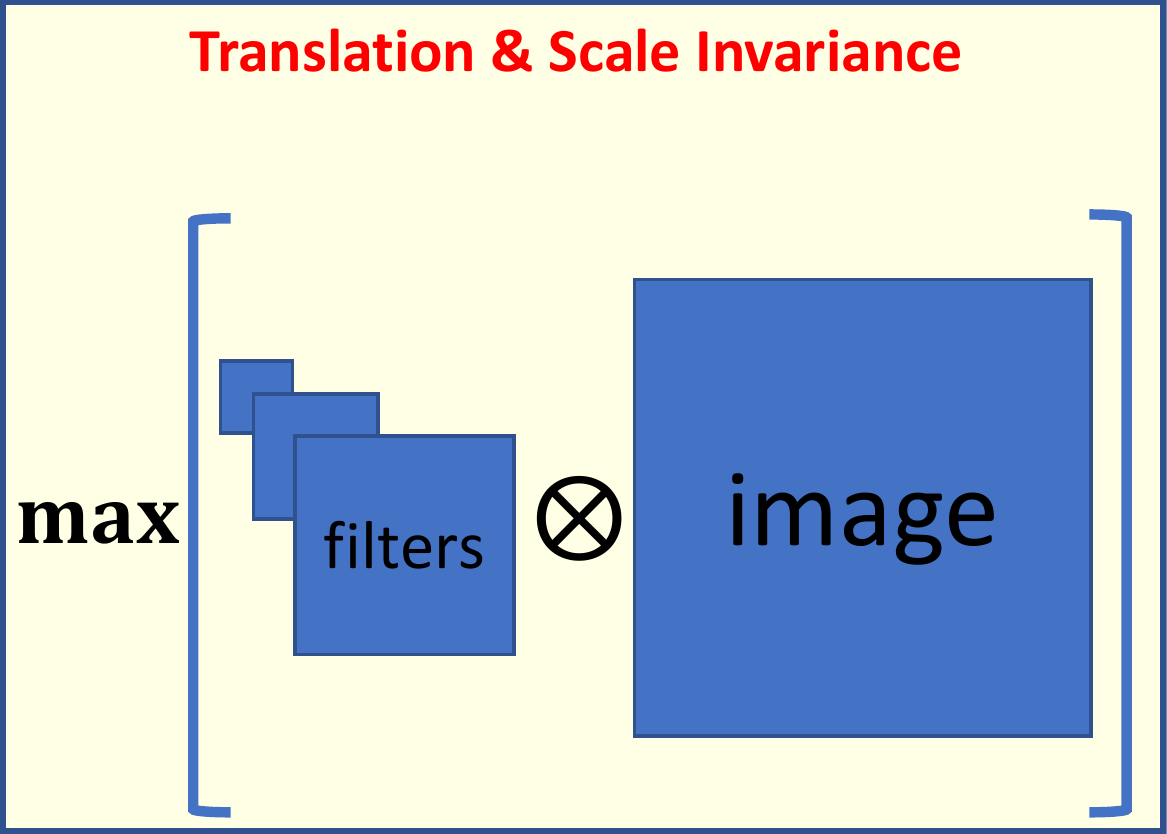}     
     \centerline{\footnotesize (a) Multi-scale maxout block}
   \end{minipage}\hfill
   \begin{minipage}{0.495\linewidth}
     \centering
     \includegraphics[width=.9\linewidth]{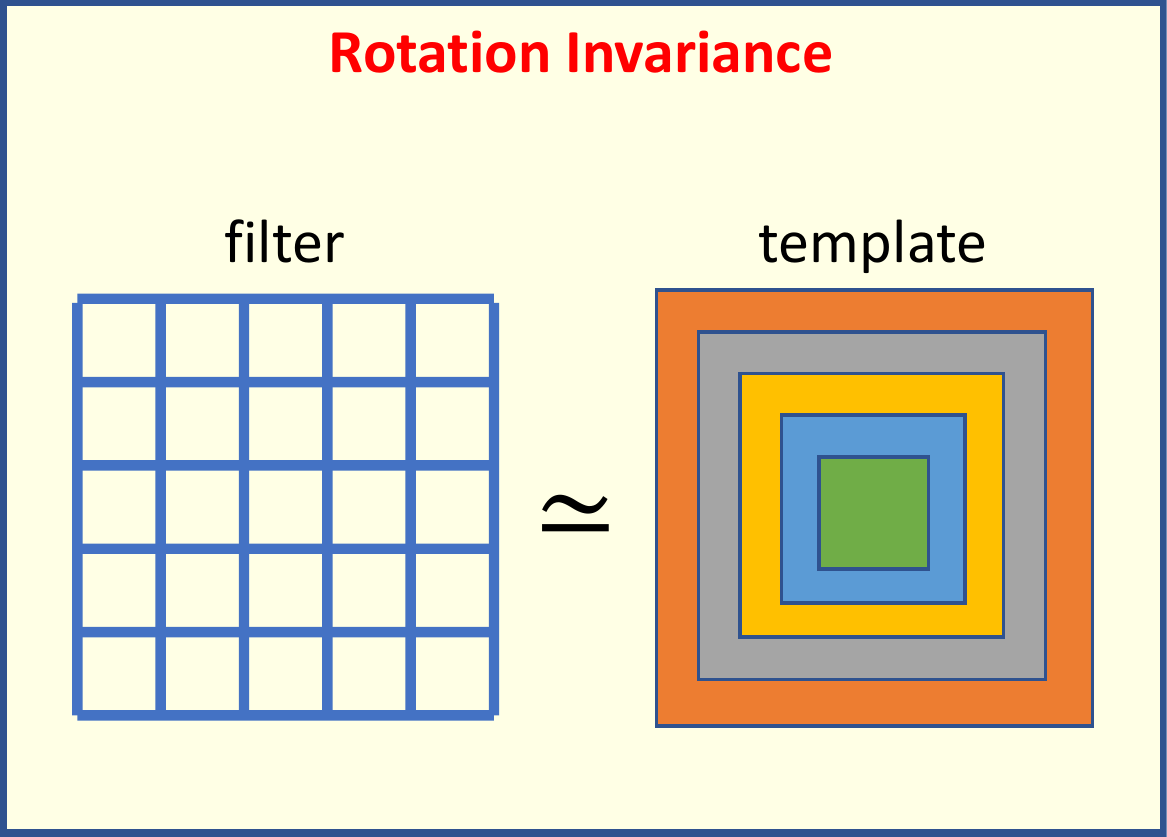}     
     \centerline{\footnotesize (b) Regularization}     
   \end{minipage}
   \vspace{2mm}
   \caption{\footnotesize To learn affine-invariant representations, we propose {\bf (a)} a multi-scale maxout convolutional network block to handle translation and scale, and {\bf (b)} a regularizer to handle rotation. We use (a) for constructing our network, and embed (b) into our learning.}\label{fig:blocks}
   \vspace{-3mm}
\end{figure}

\section{Our Approach}
\bfsection{Overview}
To achieve translation and scale invariance, we propose a multi-scale maxout block as shown in Fig. \ref{fig:blocks}(a), a set of filters with different predefined sizes are applied to images with convolution, and then the maxout operator is used to locate the maximum response per pixel among the filters. 
Mathematically this block can be formulated as
\begin{align}
    \max_{\omega\in\Omega}\left\{\omega\otimes\mathbf{I}_{ij}\right\},  \forall (i,j),
\end{align}
where $\otimes$ denotes the convolution operator, $\omega\in\Omega$ denotes a 2D spatial filter, $\mathbf{I}$ denotes an image, and $\omega\otimes\mathbf{I}_{ij}$ denotes the scalar output of the convolution at pixel $(i,j)$.

In contrast to rotation-invariant networks such as RotEqNet, there is no rotation constraint on the design of network architectures including filters. Instead, we impose such constraint on learning with our rotation-invariant regularizer. Similar to other regularizers, ours encodes the prior knowledge of filters that we would like to learn (denoted as the template in Fig. \ref{fig:blocks}(b)). Inspired by Harmonic Networks, ideally the learned filters should be symmetric along all possible directions, like circles. Due to the discretization of images, however, we propose an alternative to represent such symmetry that can be learned efficiently and effectively.

\bfsection{Learning Problem}
In this paper we consider the following optimization problem:
\begin{align}\label{eqn:learning_problem}
    \min_{\omega\in\Omega, \theta\in\Theta}\sum_i\ell\Big(y_i, \phi(x_i, \omega)\Big) + \lambda_1\mathcal{R}_1(\omega) + \lambda_2\mathcal{R}_2(\omega, \theta),
\end{align}
where $\{x_i, y_i\}\subseteq\mathcal{X}\times\mathcal{Y}$ denotes the training data with image $x_i\in\mathcal{X}, \forall i$ and its class label $y_i\in\mathcal{Y}$, $\omega\in\Omega$ denotes the parameters for the network defined by function $\phi:\mathcal{X}\times\Omega\rightarrow\mathcal{Y}$, $\theta\in\Theta$ denotes the templates in the feasible space $\Theta$ that $\omega$ should match with, $\ell:\mathcal{Y}\times\mathcal{Y}\rightarrow\mathbb{R}$ denotes the loss function, $\mathcal{R}_1$ denotes the weight decay with $\ell_2$ norm, $\mathcal{R}_2:\Omega\times\Theta\rightarrow\mathbb{R}$ denotes the regularizer that measures the difference between $\omega$ and $\theta$, and $\lambda_1, \lambda_2\geq0$ are predefined constants. Different from conventional CNNs, here we propose learning not only the network weights $\omega$ but also the matching templates $\theta$ within the feasible space $\Theta$ that encodes certain constraints on the templates such as symmetry. 
In the sequel we will explain how to effectively design a scale-invariant network $\phi$, and how to efficiently construct a rotation-invariant regularizer $\mathcal{R}_2$.

\subsection{Network Architecture}\label{ssec:architecture}
\begin{figure}[t]
    \centering
    \includegraphics[width=.85\linewidth]{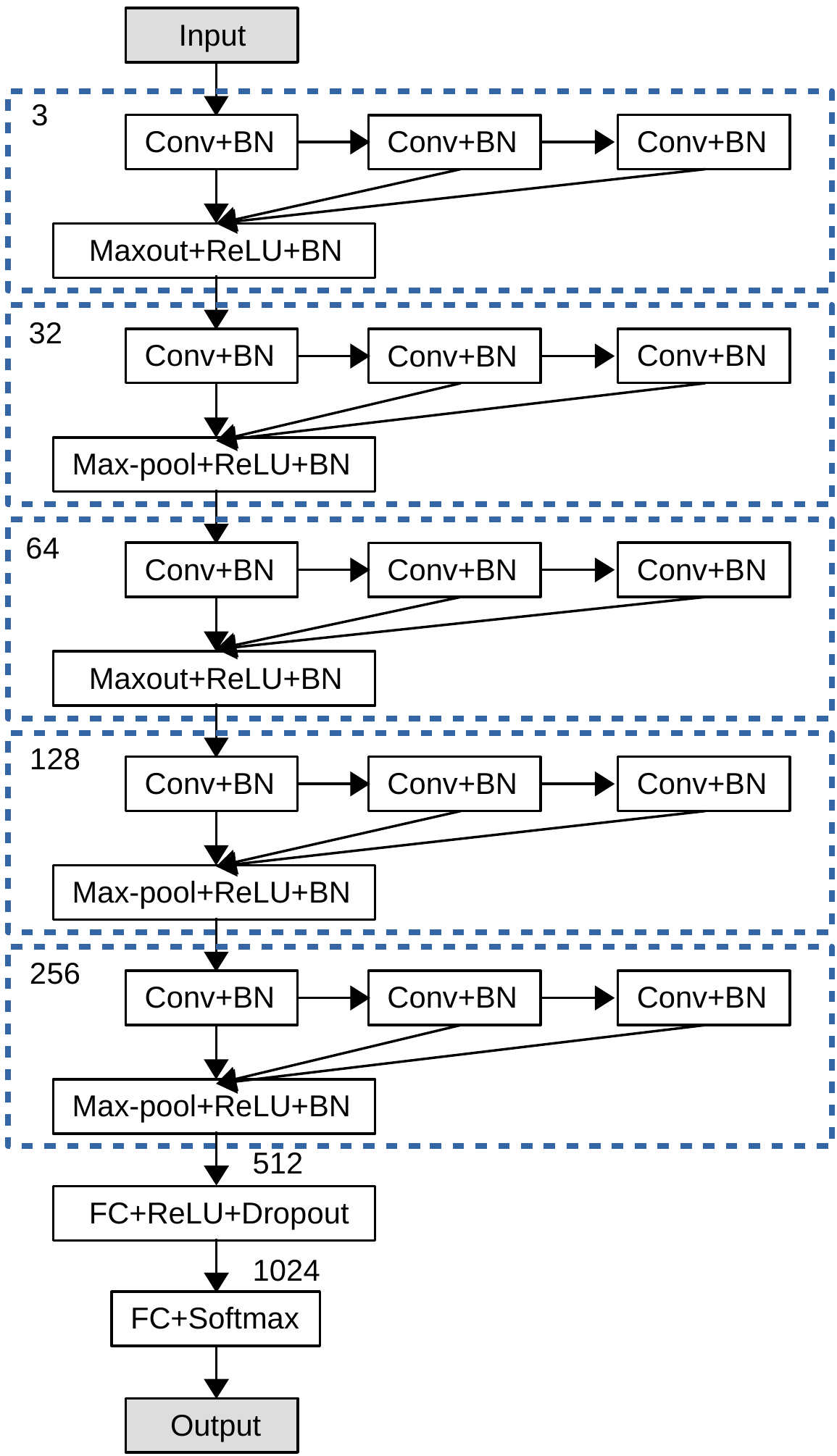}     
    \vspace{2mm}
    \caption{\footnotesize Illustration of the network we use in our experiments for learning affine-invariant features. Each dashed block is a multi-scale maxout block accounting for scale invariance, and the numbers here denote the default dimensions of inputs for the corresponding blocks and layers.}\label{fig:pi-net}
\end{figure}

We illustrate our network in Fig. \ref{fig:pi-net}, where all the operations are basic and widely used in CNNs such as batch normalization (BN) \cite{ioffe2015batch}, and ``+'' denotes one operation followed by the other. 
Due to the small image sizes (\eg $32\times32$ pixels) in our experiments, we conduct downsampling for three times only using max-pooling. In each block the first convolutional layer is responsible for mapping the inputs into a higher dimensional space, \eg $3\rightarrow32$, and the other two convolutional layers learn the (linear) transformation in the same space, \eg $32\rightarrow32$. For grayscale images, the input dimension is changed from 3 to 1.

Different from existing networks such as GoogLeNet and TI-Pooling, we propose extracting features within different scales using a sequence of convolutional operations. Considering the trade-off between computational efficiency and accuracy, we only exploit three scales, \ie $3\times3, 5\times5, 7\times7$, using fixed filter size of $3\times3$ in each convolutional layer, and use maxout to select a scale with the maximum response. This scale is taken as the best one to fit for the object. In fact we use two and three $3\times3$ convolutions to approximate the responses with filter sizes of $5\times5$ and $7\times7$, respectively, for efficient computation. With the increase of the network depth, information within larger scales (\ie receptive field) can be extracted as well. 

We also find that the network depth is more important than the network width \wrt the accuracy. It has been demonstrated in Wide Residual Networks (WRN) \cite{zagoruyko2016wide} that wider networks can improve the performance. In contrast to the parallel mechanism in WRN, in each block we apply convolutions sequentially. Note that the proposed mechanism can be integrated with other networks as well.

\subsection{Training with Rotation-Invariant Regularizer}
\subsubsection{General Formulation}
As illustrated in Fig. \ref{fig:blocks}(b), in order to enforce the filters to satisfy certain spatial properties such as rotation invariance, the templates here need to be constructed in certain way to encode such properties. Therefore, we propose the following general formulation for rotation-invariant regularizers:
\begin{align}
    & \mathcal{R}_2(\omega,\theta) \\
    & = \mathbb{E}_{k\sim\mathcal{K}}\left[\sum_{m=-p_k}^{p_k}\sum_{n=-q_k}^{q_k} d\Big(\omega_k(m,n), \theta_k(h(m,n))\Big)\right], \nonumber 
\end{align}
where $k\in\mathcal{K}$ denotes the index of a 2D spatial filter, 
$\mathbb{E}_{k\sim\mathcal{K}}$ denotes the expectation over all 2D spatial filters, $(m,n)$ denotes the 2D-index of a weight in the $k$-th filter with size $(M_k,N_k)$, $p_k=\left\lceil\frac{M_k}{2}\right\rceil, q_k=\left\lceil\frac{N_k}{2}\right\rceil$, $\lceil\cdot\rceil$ denotes the ceiling function, $d:\mathbb{R}\times\mathbb{R}\rightarrow\mathbb{R}$ denotes a distance function, $h:\mathbb{R}\times\mathbb{R}\rightarrow\mathbb{R}$ denotes a hash function that determines the weight pattern in the templates for matching, and correspondingly $\theta:\mathbb{R}\rightarrow\mathbb{R}$ is a learnable function.

\bfsection{Choices of Distance Function $d$}
In general we do not have any explicit requirement on $d$. For instance, it can be $\ell_1$-norm, $\ell_2$-norm, or group sparsity norm such as $\ell_{2,1}$-norm. Moreover, this distance measure can be conducted in not only Euclidean but also non-Euclidean spaces such as manifold regularization \cite{belkin2006manifold}, which will be appreciated in geometric deep learning \cite{bronstein2017geometric}.

\begin{figure}[t]
   \begin{minipage}{0.495\linewidth}
     \centering
     \includegraphics[width=.5\linewidth]{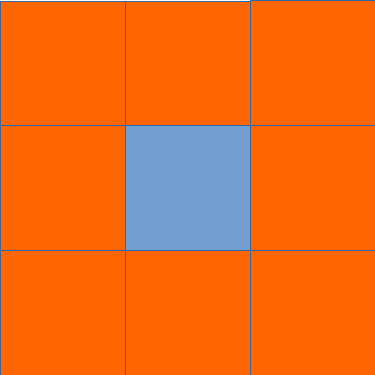}
     \centerline{\footnotesize (a)}
   \end{minipage}\hfill
   \begin{minipage}{0.495\linewidth}
     \centering
     \includegraphics[width=.5\linewidth]{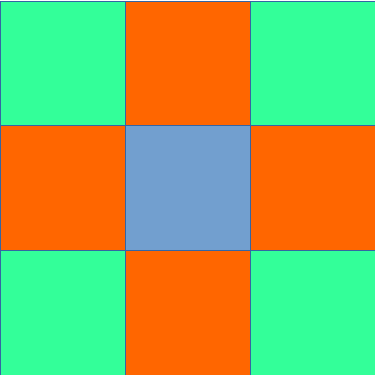}
     \centerline{\footnotesize (b)}     
   \end{minipage}
   \vspace{2mm}
   \caption{\footnotesize Examples of weight patterns, defined by hash function $h$, that can be used to approximate circular patterns for rotation invariance. In each subfigure the same color denotes the same weight.}
   \label{fig:patterns}
   \vspace{-4mm}
\end{figure}

\bfsection{Choices of Hash Function $h$}
For rotation invariance, ideally it should be a circular pattern defined by $h(m,n)=(m^2+n^2)^{\frac{1}{2}}$ in a continuous space. Due to the discretization of images, however, it hardly forms circles in filters without interpolation which will significantly increase the computational complexity in convolution. Instead, we propose learning some simpler patterns that can be used to approximate circles. For instance, we illustrate two exemplar patterns for filters with size $3\times3$ in Fig. \ref{fig:patterns}, where the patterns in (a) and (b) are defined by $h(m,n)=\left\lfloor(m^2+n^2)^{\frac{1}{2}}\right\rfloor$ and $h(m,n)=\left\lceil(m^2+n^2)^{\frac{1}{2}}\right\rceil$, respectively, and $\lfloor\cdot\rfloor$ is the floor function. Other hash functions may be also applicable here, but finding the best one is outside the scope of this paper.

\subsubsection{An Empirical Showcase}
In this section we will show a specific regularizer that we use in our experiments later. For the simplicity and efficiency, we decide to employ the least square loss for $d$ and the pattern in Fig. \ref{fig:patterns}(a) for $h$ without fine-tuning the accuracy on the data sets.

Specifically we define our empirical rotation-invariant regularizer as follows:
\begin{align}\label{eqn:R_rot}
    & \mathcal{R}_2(\omega,\theta) \\ 
    & = \mathbb{E}_{k\sim\mathcal{K}}\left[\sum_{m,n\neq0} \left(\omega_k(m,n) - \frac{\sum_{m',n'\neq0}\omega_k(m',n')}{p_kq_k-1}\right)^2\right], \nonumber 
\end{align}
where $\theta_k(h(m,n))=\frac{\sum_{m',n'\neq0}\omega_k(m',n')}{p_kq_k-1}$ is a scalar. 

Similar to the center loss in \cite{wen2016discriminative}, here we aims to reduce the variance among the weights in each 2D spatial filter with $3\times3$ pixels, on average. Meanwhile, the patterns in the templates are updated automatically with the mean of the weights. In this way we can learn filters that can better approximate 2D spatial circular patterns for rotation invariance. In backpropagation, since $\mathcal{R}_2(\omega,\theta)$ in Eq. \ref{eqn:R_rot} is always differentiable \wrt $\omega_k, \forall k$, any deep learning solver such as stochastic gradient descent (SGD) can be used to train the network with our rotation-invariant regularizer.

\bfsection{Discussion}
Recall that Fig. \ref{fig:patterns} essentially encodes the structural patterns that we expect for learned filters to handle rotation. One may argue that we can enforce such structures into learning strictly by converting the regularizer $\mathcal{R}_2$ in Eq.~\ref{eqn:learning_problem} into constraints and solving a constrained nonconvex optimization problem. We decide not to do so because potentially the new problem will be much harder to be solved than the one in Eq.~\ref{eqn:learning_problem}. Besides since the structures in Fig.~\ref{fig:patterns} are already the approximation of the circular structure, we do not necessarily guarantee that all the weights with the same color are identical. More freedom as in regularization may lead to a compensation for the loss of the structural approximation in terms of accuracy.

\section{Experiments}\label{sec:exp}

\setlength{\tabcolsep}{1.1pt}
\begin{table*}\small
    \caption{\footnotesize Test accuracy (\%) comparison on different datasets under two training settings: {\bf (F)} with all the images, and {\bf (10)} with 10 random images per class.}
    \vspace{-2mm}
	\begin{center}    
		\begin{tabular}{|c||c|c|c|c|c|c|c|c|c|c|}
			\hline & {\bf Ours} & RotEqNet & Harmonics & TI-Pooling & GCNN & STN & ResNet-32 & CapsNet & GoogLeNet & DCN \\ 
            \hline aff. (F) & {\bf 99.08} & 94.81& 94.20& 94.72& 95.43& 98.24& 95.76& 97.30& 98.12&87.70 \\
            \hline rot (F) & {\bf 98.92} & 98.91& 98.31& 98.80& 97.72& 97.12& 95.96& 96.73& 98.29&92.69 \\
            \hline T. S. (F) &  {\bf 98.87} & 94.79& 94.02&97.47 &91.47 &40.87 & 88.35& 95.15& 91.16&68.29 \\
            \hline Ave. (F) & {\bf 98.95} &96.17 &95.51 &96.99 & 94.87&78.74 &93.36 &96.39 &95.85 &82.75 \\
			\hline
			\hline aff. (10) & {\bf 85.06$\pm$0.90} & 45.91$\pm$3.85 & 56.41$\pm$3.66 & 34.40$\pm$1.54 & 25.67$\pm$1.99 & 23.85$\pm$0.12 & 18.56$\pm$0.27 & 19.74$\pm$0.39 &50.77$\pm$0.20 &10.74$\pm$0.52 \\
            \hline rot (10) & {\bf 87.49$\pm$0.56} & 84.18$\pm$2.17 &54.67$\pm$2.65 &83.86$\pm$0.88 &45.12$\pm$2.48 &66.72$\pm$0.72 &49.31$\pm$0.30 &81.17$\pm$0.17 &82.20$\pm$0.35 &49.69$\pm$0.33   \\
            \hline T. S. (10) & {\bf 84.15$\pm$0.48} &26.43$\pm$0.85 &27.57$\pm$0.94 &47.31$\pm$1.37 &29.66$\pm$0.63 &27.72$\pm$1.74 &28.20$\pm$2.06 &54.35$\pm$0.56  &32.49$\pm$0.29 &28.52$\pm$0.26 \\
            \hline Ave. (10) & {\bf 85.56}  &52.17 &46.21 &55.19 &33.48 & 39.43&32.02 &51.75 &55.15 & 29.65\\
			\hline
		\end{tabular}
	\end{center}
    \vspace{-7mm}
	\label{tab:test_acc}
\end{table*}

\begin{figure*}[t]
   \begin{minipage}{.325\linewidth}
     \centering
     \includegraphics[width=1.05\linewidth]{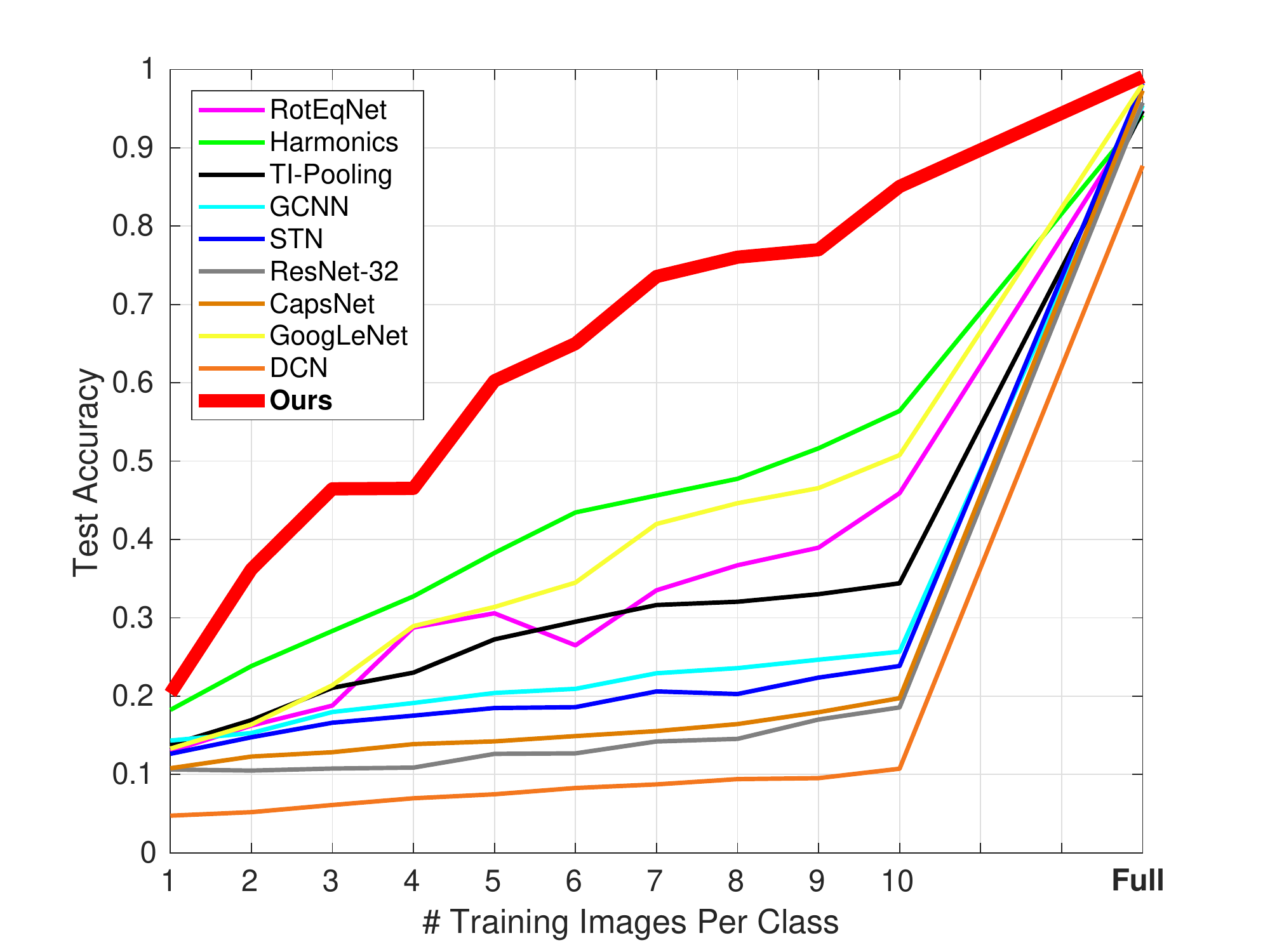}
     \centerline{\footnotesize (a) affNIST}
   \end{minipage}\hfill
   \begin{minipage}{.325\linewidth}
     \centering
     \includegraphics[width=1.05\linewidth]{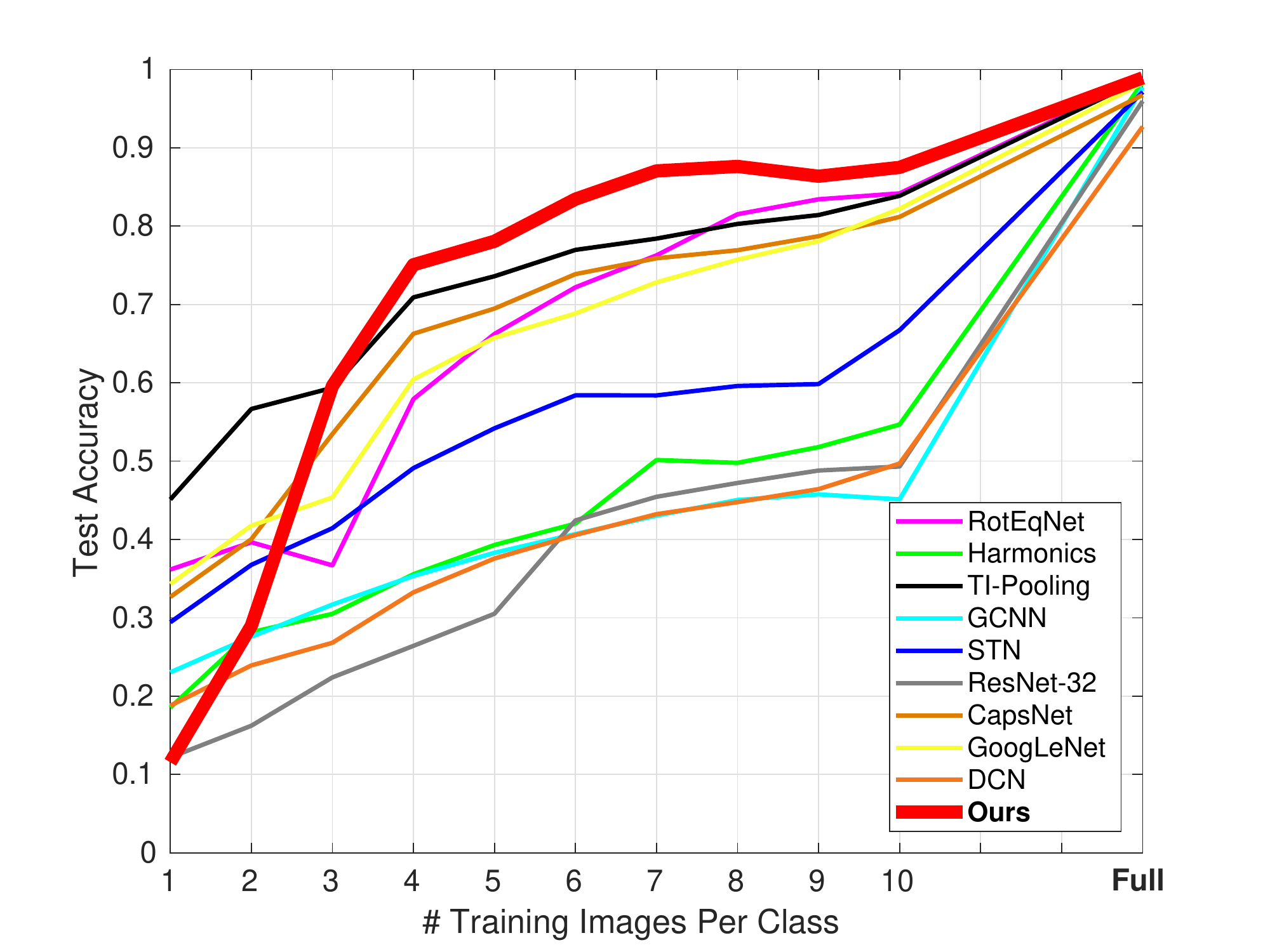}
     \centerline{\footnotesize (b) MNIST-rot}
   \end{minipage}
   \begin{minipage}{.325\linewidth}
     \centering
     \includegraphics[width=1.05\linewidth]{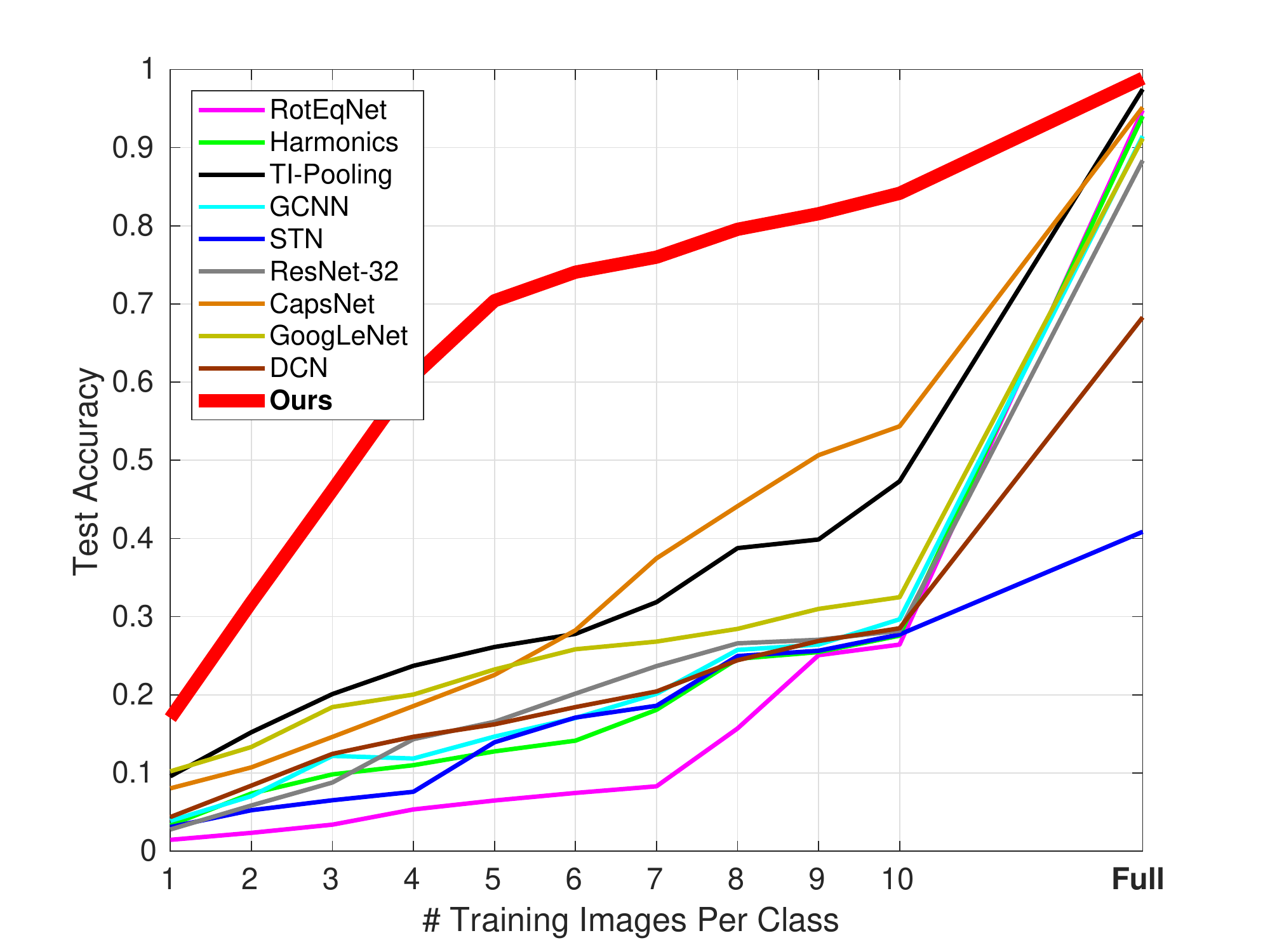}
     \centerline{\footnotesize (c) Traffic Sign}     
   \end{minipage}
   \caption{\footnotesize Test accuracy comparison of different networks on the three data sets. ``Full'' here indicates that we use all the training images. Our approach significantly outperforms the state-of-the-art, especially with small numbers of training images.}
   \label{fig:test_acc}
  \vspace{-3mm}
\end{figure*}

\begin{figure}[t]
      \centering
      \includegraphics[width=.7\linewidth]{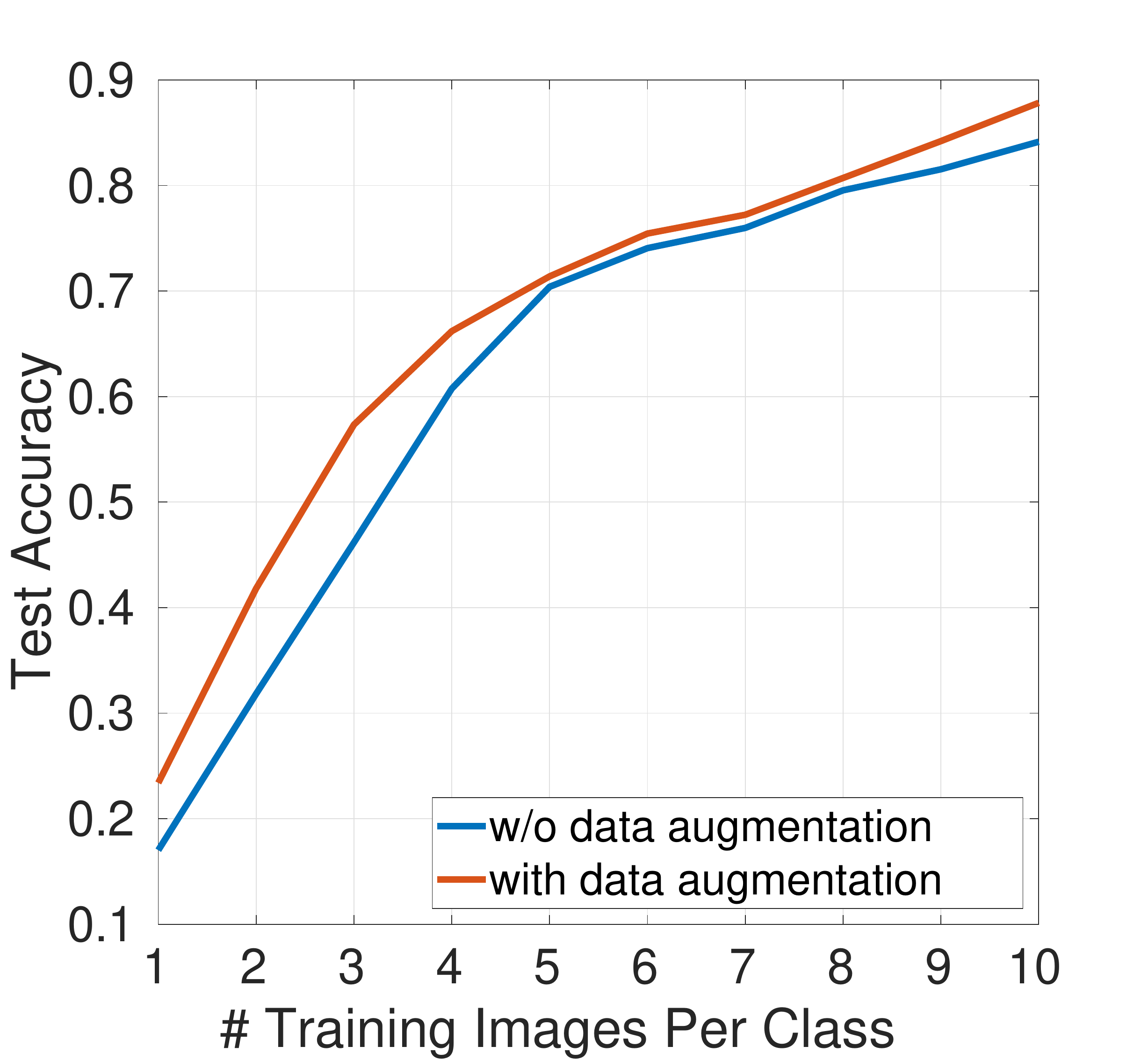}
      \caption{\footnotesize Data augmentation comparison on Traffic Sign.}
      \label{fig:data_augmentation}
      \vspace{-5mm}
\end{figure}

\begin{figure*}[t]
   \begin{minipage}{.495\linewidth}
     \centering
     \includegraphics[width=.95\linewidth]{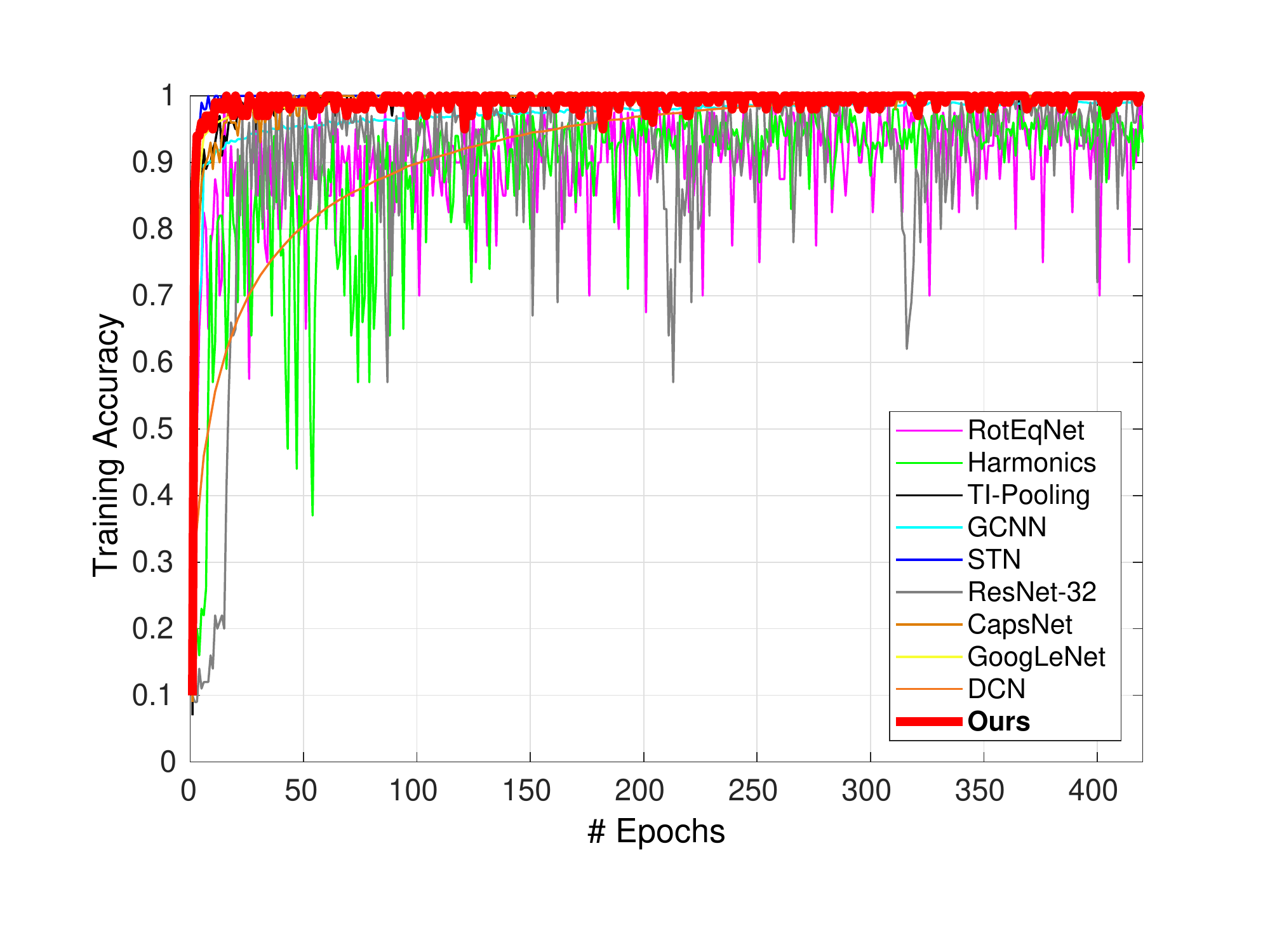}
   \end{minipage}\hfill
   \begin{minipage}{.495\linewidth}
     \centering
     \includegraphics[width=.95\linewidth]{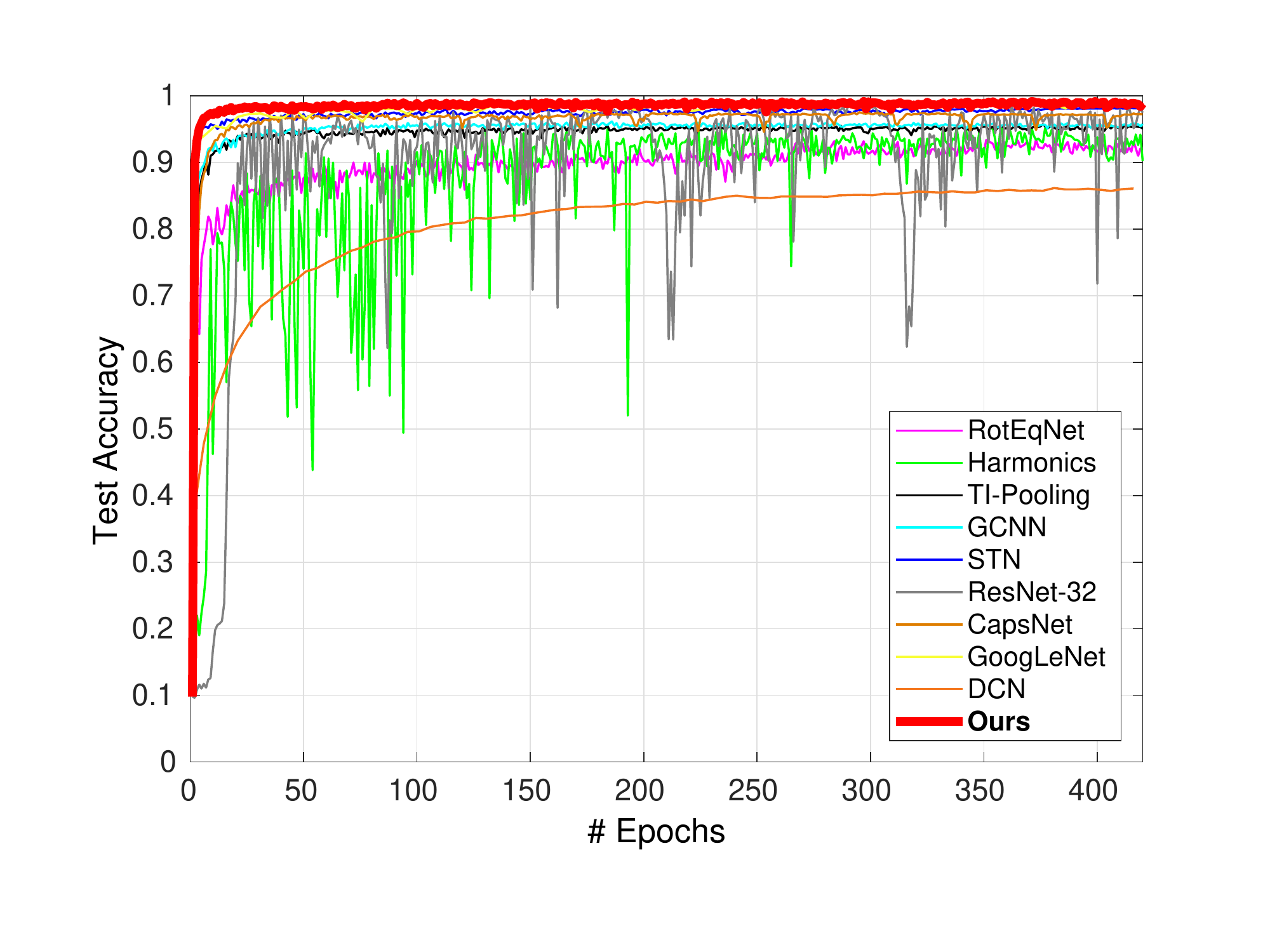}
   \end{minipage}
   \vspace{-3mm}
   \caption{\footnotesize Illustration of training/testing behavior of different networks on affNIST.}
   \label{fig:train_test_curves}
  \vspace{-3mm}
\end{figure*}

\subsection{Benchmark Data with Affine Transformations}
\subsubsection{Experimental Setup}

\bfsection{Data Sets}
We test our approach on three benchmark data sets, affNIST \cite{sabour2017dynamic}, MNIST-rot \cite{larochelle2007empirical}, and Traffic Sign \cite{Stallkamp2011}. 

affNIST is created by applying random small affine transformations 
to each $28 \times 28$ grayscale image in MNIST \cite{10027939599} (10 classes). It is designed for testing the tolerance of an algorithm to such transformations. 
There are 60K training and validation samples and 10K test samples in affNIST with size $40\times40$ pixels. To facilitate the data processing in training, we resize all the images to $32 \times 32$ pixels.

MNIST-rot \cite{larochelle2007empirical} is another variant of MNIST, where a random rotation between $0^\circ$ and $360^\circ$ is applied to each image. It has 10K/2K/50K training/validation/test samples. To facilitate the data processing in training, we again resize all the grayscale images to $32 \times 32$ pixels.

Traffic Sign contains 43 classes with unbalanced class frequencies, 34799 training RGB images, and 12630 testing RGB images with size of $32\times32$ pixels. It reflects the strong variations in visual appearance of signs due to distance, illumination, weather conditions, partial occlusions, and rotations, leading to a very challenging recognition problem.

\bfsection{Networks}
We compare our approach with some state-of-the-art networks with similar model complexity to ours, \ie RotEqNet \cite{marcos2017rotation}\footnote{\url{https://github.com/COGMAR/RotEqNet}}, Harmonics \cite{worrall2017harmonic}\footnote{\url{https://github.com/deworrall92/harmonicConvolutions}}, TI-Pooling \cite{laptev2016ti}\footnote{\url{https://github.com/dlaptev/TI-pooling}}, GCNN \cite{cohen2016group}\footnote{\url{https://github.com/tscohen/gconv_experiments}}, STN \cite{jaderberg2015spatial}\footnote{\url{https://github.com/kevinzakka/spatial-transformer-network}}, ResNet-32 \cite{he2016deep}\footnote{\url{https://github.com/tensorflow/models/tree/master/research/resnet}}, CapsNet \cite{sabour2017dynamic}\footnote{\url{https://github.com/naturomics/CapsNet-Tensorflow}}, GoogLeNet \cite{szegedy2015going}\footnote{\url{https://github.com/flyyufelix/cnn_finetune/blob/master/googlenet}}, and DCN \cite{dai2017deformable}\footnote{\url{https://github.com/felixlaumon/deform-conv}}. Specifically TI-Pooling is designed for scale invariance, RotEqNet, Harmonics, and GCNN are designed for rotation invariance. We use the public code for our comparison.

We implement our default network using Tensorflow and following the architecture in Fig. \ref{fig:pi-net} with the default numbers of channels. 
Note that the implementation of the networks in our comparison may be different, (\ie
GCNN$\rightarrow$Chainer; GoogLeNet, DCN$\rightarrow$Keras; CapsNet, TI-Pooling, Harmonics, STN, ResNet-32$\rightarrow$Tensorflow; RotEqNet$\rightarrow$Pytorch) which may lead to various computational efficiency.

\bfsection{Training Protocols}
We tune each network to report its best performance on the data sets. By default we train the networks for 42000 iterations with mini-batch size 100, weight decay $\lambda_1=0.0005$, and momentum 0.9. The global learning rate is set to 0.01 or 0.0001 when trained using all or a few training images per class, respectively, and it is reduced by 0.1 twice at the 20000 iteration and the 30000 iteration as well. For each network the hyper-parameter tuning starts with the default setting, and the best setting may be slightly different from the default. We follow this default setting in all the experiments and set $\lambda_2=150$. The numbers reported here are the average over three trials.

To do fair comparison, we follow the settings for data augmentation in the publications of most of the competitors. Specifically, by default on affNIST and Traffic Sign we do not employ data augmentation, but on MNIST-rot we do.

\begin{figure}[t]
      \vspace{2mm}
      \centering
      \includegraphics[width=\linewidth]{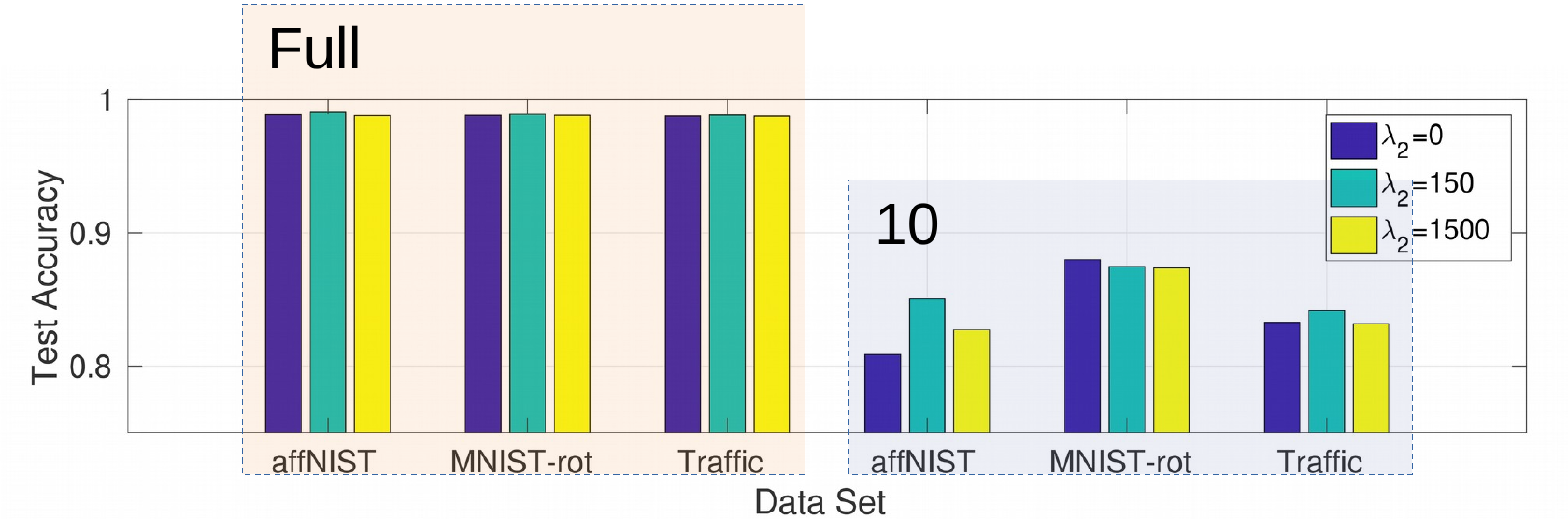}
      \caption{\footnotesize Illustration of the effect of $\lambda_2$ in Eq. \ref{eqn:learning_problem} on test accuracy.}
      \label{fig:lambda}
      \vspace{-3mm}
\end{figure}

\begin{figure*}[t]
   \begin{minipage}{.325\linewidth}
     \centering
     \includegraphics[width=1.1\linewidth]{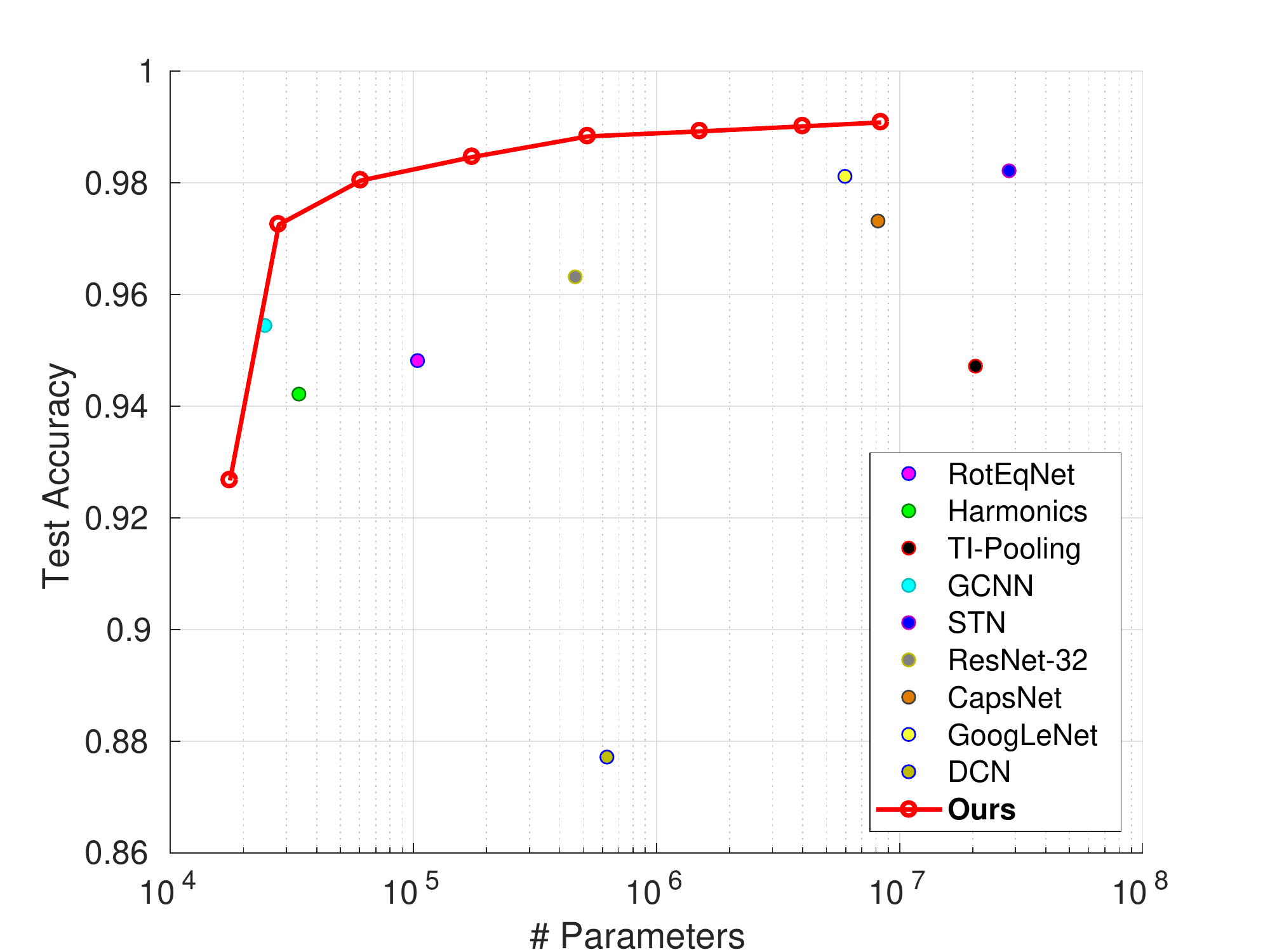}
     \centerline{\footnotesize (a) affNIST}
   \end{minipage}\hfill
   \begin{minipage}{.325\linewidth}
     \centering
     \includegraphics[width=1.1\linewidth]{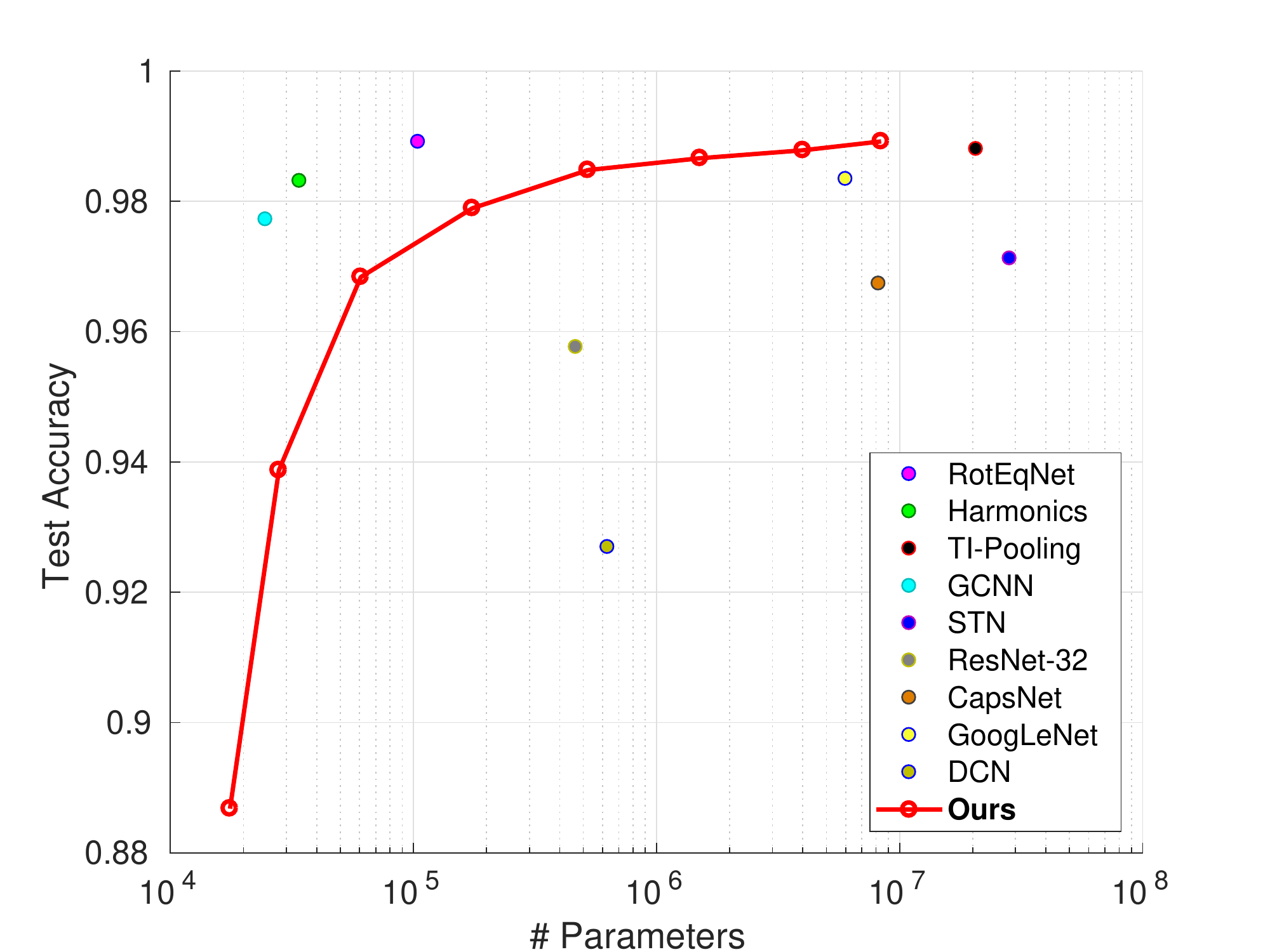}
     \centerline{\footnotesize (b) MNIST-rot}     
   \end{minipage}
   \begin{minipage}{.325\linewidth}
     \centering
     \includegraphics[width=1.1\linewidth]{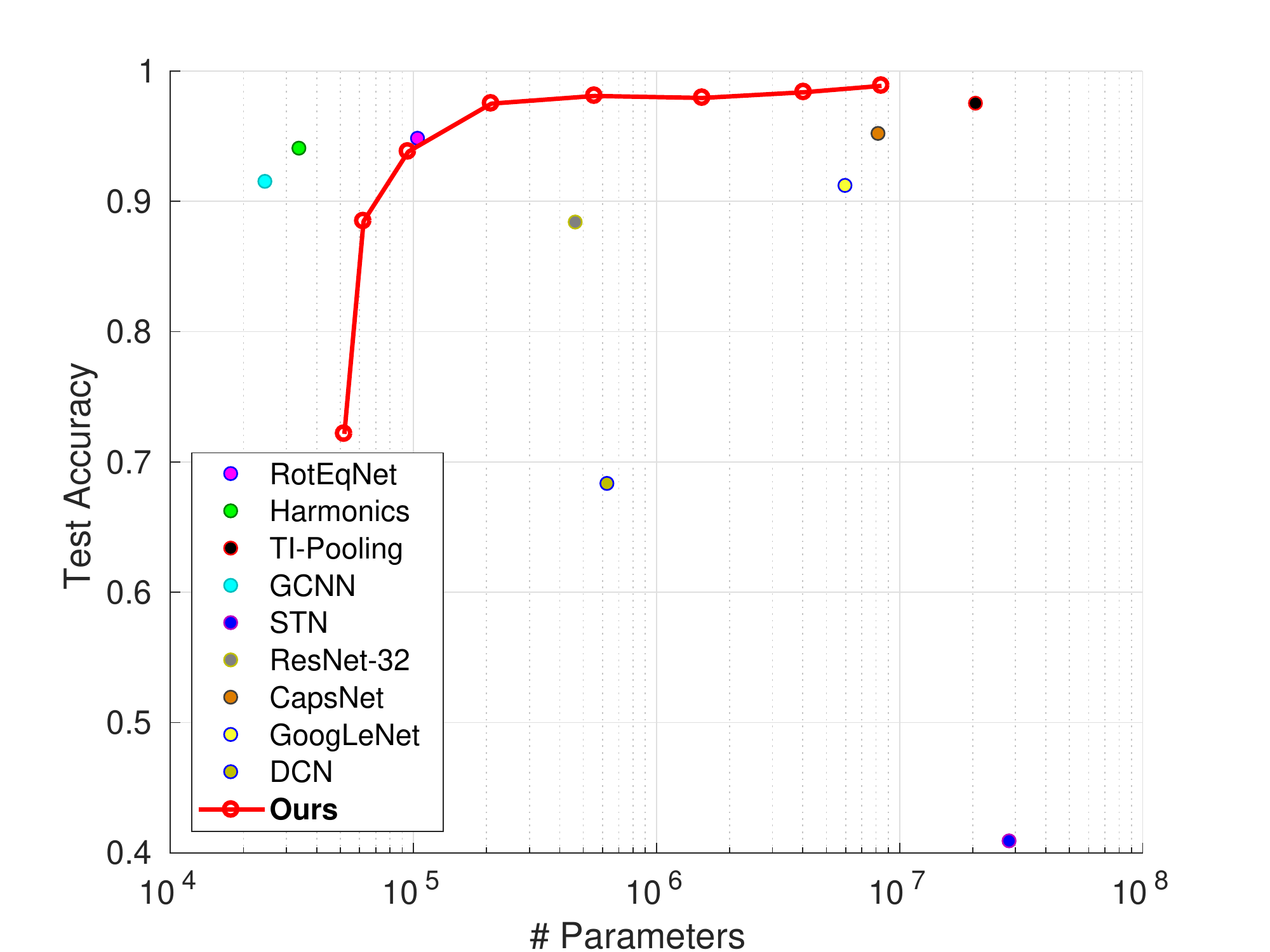}
     \centerline{\footnotesize (c) Traffic Sign}     
   \end{minipage}
   \vspace{2mm}
   \caption{\footnotesize Test accuracy comparison with the others using different numbers of parameters. Best viewed in color.}
   \label{fig:test_acc_vs_num_param}
  \vspace{-3mm}
\end{figure*}

\begin{figure*}[t]
   \begin{minipage}{.495\linewidth}
     \centering
     \includegraphics[width=.95\linewidth]{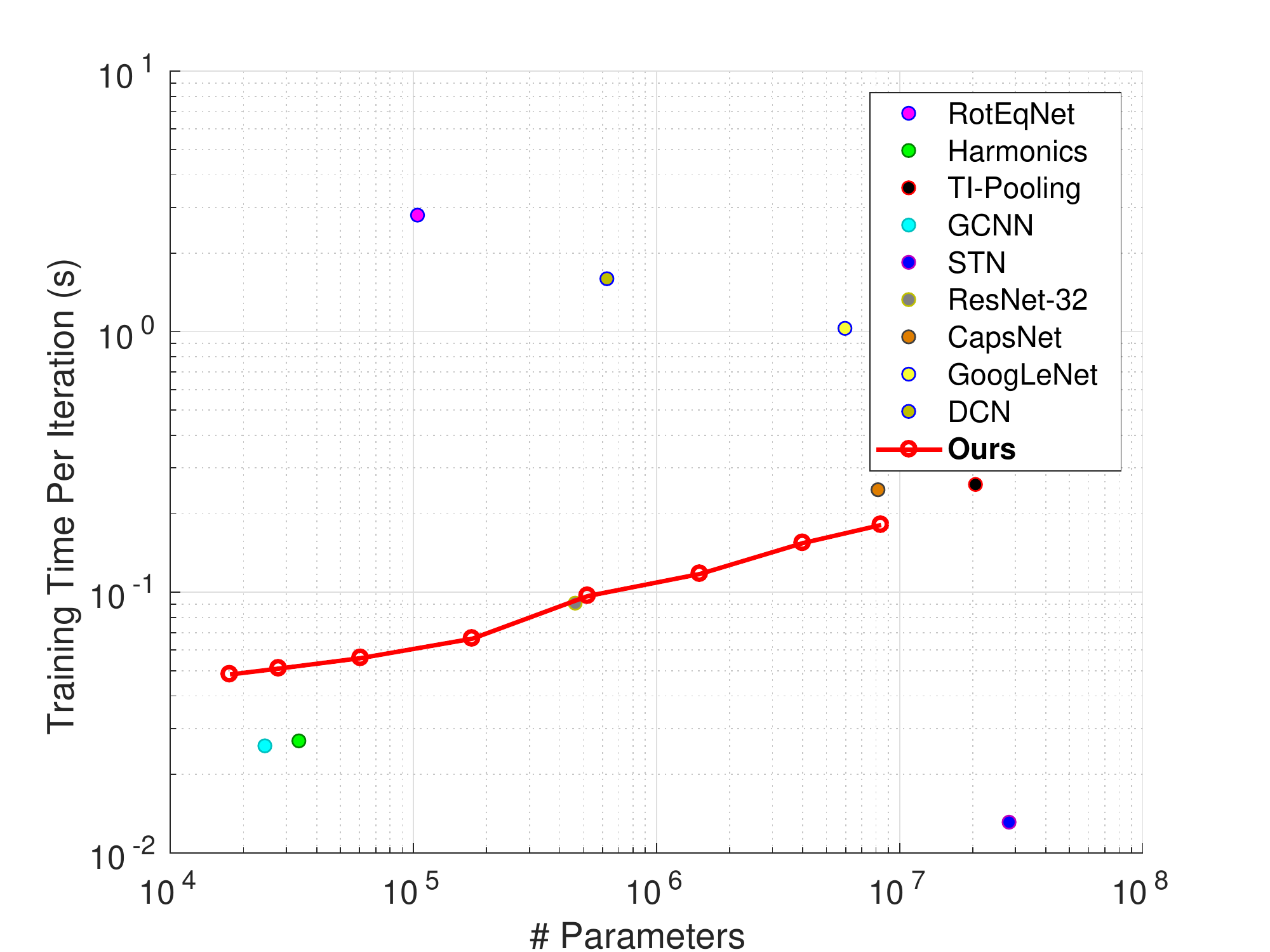}
   \end{minipage}\hfill
   \begin{minipage}{.495\linewidth}
     \centering
     \includegraphics[width=.95\linewidth]{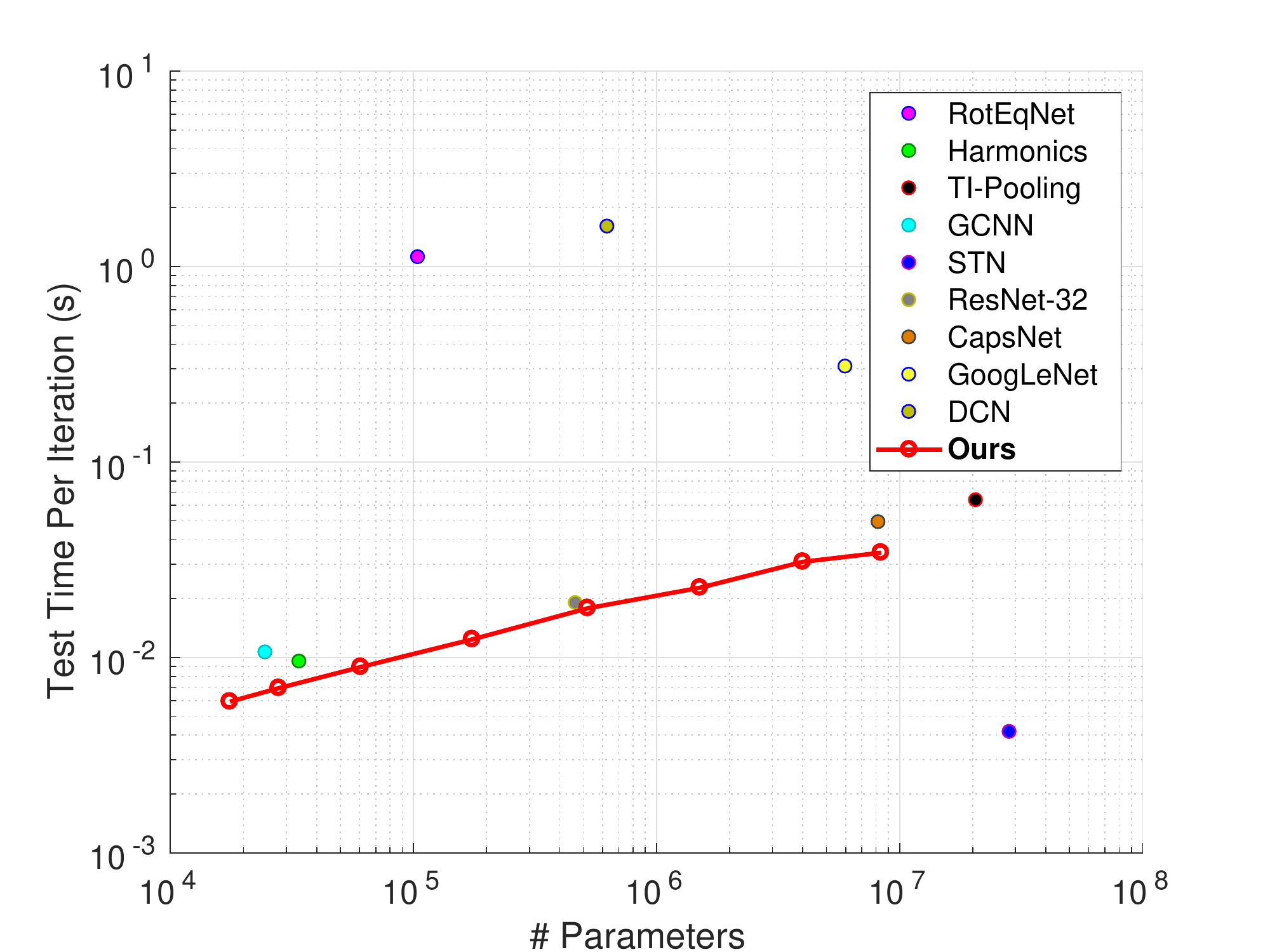}
   \end{minipage}
   \vspace{2mm}
   \caption{\footnotesize Training/Test time comparison with the others using different numbers of parameters. Best viewed in color.}
   \label{fig:train_test_time}
  \vspace{-3mm}
\end{figure*}

\subsubsection{Results}

\bfsection{Better Generalization, Data-Efficiency, \& Robustness}
We summarize the test accuracy comparison in Table \ref{tab:test_acc}. As we see, using either all or 10 random training/validation images per class, our method consistently outperforms the competitors on the three data sets with a margin of {\bf 1.96\%} or {\bf 30.37\%}. Using the full set the stds of all the methods are small and similar, and thus we do not show the numbers.

To better demonstrate the data-efficiency, we illustrate test accuracy comparison using few random training/validation images per class in Fig. \ref{fig:test_acc}. Overall, our method works significantly better than the competitors with large margins. Note that on MNIST-rot our performance is worse than some of the competitors when using 1 or 2 images per class for training. A possible reason may come from data augmentation. Another reason is that some of the networks are designed specifically for rotation invariance and this data set just fits for this purpose. With the increase of the numbers of training samples, however, our method again beats all the competitors. It is worth mentioning that in Harmonics Networks \cite{worrall2017harmonic}, similar experiments on MNIST-rot were conducted to show data-efficiency and robustness of the approach. Using $\frac{1}{6}$ of the full training/validation data Harmonics lost about 3\%. Here we compare different networks using less than ${\bf \frac{1}{120}}$ to show the superiority of our method over the others. Empirically we observe that our method can work very robustly with standard deviation of less than 1\%, in general.

In addition, we can further improve our performance using data augmentation. In Fig.~\ref{fig:data_augmentation} we illustrate the performance comparison on Traffic Sign with or without data augmentation. As we see, using 10 random training images per class we can achieve 87.84\% with 3.69\% improvement.

\bfsection{Training \& Testing Behavior}
We illustrate the training and test accuracy behavior of each network on affNIST with the full training set in Fig. \ref{fig:train_test_curves}. As we see all the networks are well trained with convergence. In the testing stage our network converge faster than most of the competitors with better accuracy. Similar observations can be made in training as well. We make similar observations on the other two data sets. From this perspective, we can also demonstrate that our method has better generalization.

\setlength{\tabcolsep}{2pt}
\begin{table}\small
	\caption{\footnotesize Effect on test accuracy (\%) of different multi-scale settings, where our default setting is 3$\times$[Conv+BN].}
	\vspace{-6mm}
	\begin{center}    
		\begin{tabular}{|c||c|c|c|}
			\hline & 2$\times$[Conv+BN] & 3$\times$[Conv+BN] & 4$\times$[Conv+BN] \\ 
            \hline affNIST (F) & 99.04 & 99.08 & 98.69\\
            \hline MNIST-rot (F) & 98.72 & 98.92 &98.97 \\
            \hline Traffic Sign (F) & 98.42 & 98.87 &98.42 \\
            \hline Average & 98.73 & 98.95 &98.69 \\
			\hline
		\end{tabular}
	\end{center}
    \label{tab:multi-scale}
    \vspace{-9mm}
\end{table}

\bfsection{Effect of Multi-Scale Maxout}
In Table \ref{tab:multi-scale} we list the test accuracy using different multi-scale settings, while fixing the parameter $\lambda_2=150$. As we see the changes between different settings are really marginal, which again demonstrates the good generalization and robustness of our method. Considering the trade-off between accuracy and computational efficiency, we choose 3$\times$[Conv+BN] as our default setting used in Fig. \ref{fig:pi-net}.

\bfsection{Effect of Rotation-Invariant Regularization}
We illustrate such effect in Fig. \ref{fig:lambda} while using the default multi-scale maxout setting. With different values where $\lambda_2=0$ means no our regularizer, we can see that using the full set for training our performances are almost identical. This is probably because the number of training images is sufficiently large to capture the scaling and rotation information already. Using a few training images, \eg 10 per class, the benefit of using our rotation-invariant regularizer becomes much clearer, especially on affNIST. Using $\lambda_2=150$ as default, there is 1.52\%, on average, improvement over that without our regularizer.

We also observe that our rotation-invariant regularizer can achieve very small numbers empirically. For instance, on affNIST the value is $2.94\times10^{-7}$, indicating that our learned filters are very close to the spatial circular patterns.

\bfsection{Behavior with Different Numbers of Parameters}
We reduce the number of parameters in our network by channel-wise shrinking. Specifically in ascending order of number of parameters, the corresponding network channels are set as follows: [4,4,4,4,4], [16,16,16,16,16], [32,32,32,32,32], [32,64,64,64,64], [32,64,128,128,128], [32,64,128,256,256], [32,64,128,256,512], followed by an FC of 1024 nodes and another FC for classification.

We first compare our performance using different numbers with the competitors in Fig. \ref{fig:test_acc_vs_num_param}. We can see that after about 200K parameters the improvement of our approach becomes slow, while before 200K our performance drops significantly with the decrease of numbers of parameters. In the figure 200K corresponds to the setting [32,64,64,64,64], whose performance is, or on par with, the best already.

We then compare the running time per iteration in both training and testing stages in Fig. \ref{fig:train_test_time}. We run all the code on the same machine with a Titan XP GPU. In training the running time includes the feedforward calculation and backpropagation inference (dominating training time), while in testing the running time only includes the feedforward calculation. As we see, in both training and testing our computational complexity grows exponentially, in general, with the number of parameters (note that the y-axis is in log-scale). Although some codes are written in different deep learning environments, we can still do a fair comparison with Harmonics and STN. Harmonics has fewer parameters, leading to faster backpropagation and thus shorter training time. The operations in Harmonic, however, is more complex than ours, and thus with a similar number of parameters our method is faster in testing. The operations in STN are much simpler than both Harmonics and ours, leading to faster running speed in both training and testing. Note that in order to further improve our computational efficiency, we can simply remove one Conv+BN in the multi-scale maxout block that can achieve similar accuracy (see Table \ref{tab:multi-scale}).

\subsection{Comparison on CIFAR-100 \cite{cifar100}}

\begin{figure}[t]
      \centering
      \includegraphics[width=.7\linewidth]{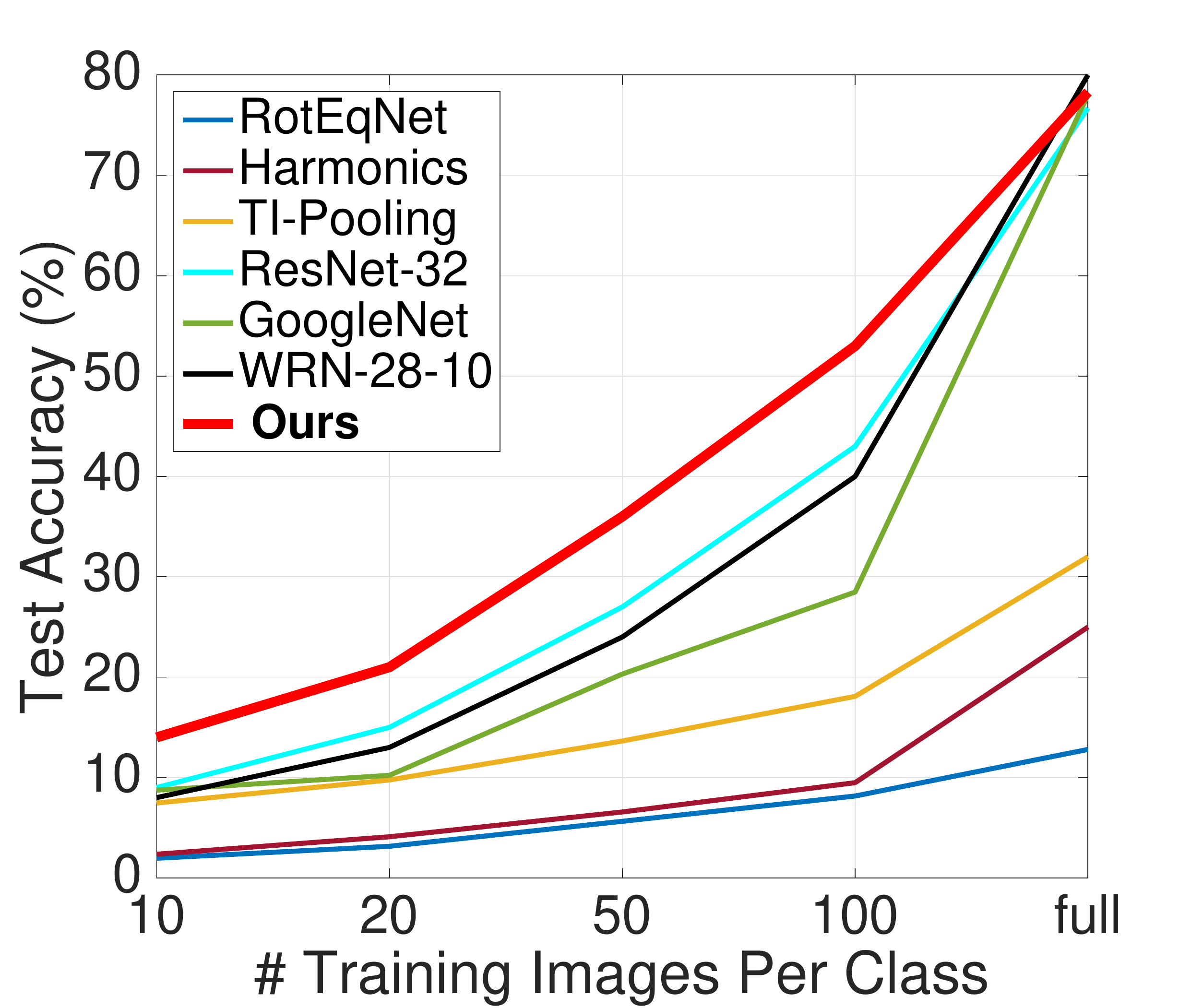}
      \caption{\footnotesize Test accuracy comparison of different networks on CIFAR-100. ``Full'' here indicates that we use all the training images. Again our approach significantly outperforms the state-of-the-art, especially with small numbers of training images.}
      \label{fig:cifar100}
      \vspace{-3mm}
\end{figure}

Beyond the benchmark data sets with affine transformations, we also test our method on ``natural'' images. For instance, we illustrate our comparison results on CIFAR-100 in Fig. \ref{fig:cifar100}. CIFAR-100 contains 60,000 $32\times32$ color images in 100 different classes, 500/100 training/testing images per class. Following the same training protocol, we randomly sample a few images per class to further demonstrate our superiority, especially on data-efficiency.

As we see in Fig. \ref{fig:cifar100}, our method significantly and consistently outperforms the competitors with a few training samples. For instance, using 100 samples per class ours achieves 52.67$\%$ test accuracy with the improvement of almost 10$\%$ over ResNet-32 (the second best). Using the full training set, ours achieves 78.33$\%$ that is slightly lower than WRN-28-10 (80.75$\%$), but higher than ResNet-32 (76.7$\%$) and GoogleNet (78.03$\%$), and dramatically higher than the other networks that learn the scale or rotation invariant representations such as TI-Pooling (31.77$\%$).

\section{Conclusion}
In this paper we propose a novel multi-scale maxout deep CNN and a novel rotation-invariant regularizer to learn affine-invariant representations for object recognition in images. Multi-scale convolution with maxout can handle translation and scale, and enforcing 2D filters to approximate circular patterns by our regularization can manage to induce invariance to rotation. By taking these as a priori knowledge, we can easily interpret our network architecture as well as its training procedure. We test our method on three benchmark data sets as well as CIFAR-100 to demonstrate its superiority over the state-of-the-art in terms of generalization, data-efficiency, and robustness. Especially, with a few training samples our method can work significantly better, leading to the hypothesis that the introduction of a priori knowledge into deep learning can effectively reduce the amount of data to accomplish the tasks. We are planning to explore more on this topic in our future work.

{\small
\bibliographystyle{ieee}
\bibliography{egbib}
}

\end{document}